\documentclass[review]{elsarticle}

\usepackage{lineno,hyperref}
\modulolinenumbers[5]
\usepackage{times}
\usepackage{epsfig}
\usepackage{graphicx}
\usepackage{amsmath}
\usepackage{amssymb}  
\usepackage{stfloats}
\usepackage{bm}
\usepackage{multirow}
\usepackage{latexsym}
\usepackage{threeparttable}
\usepackage{xcolor}
\usepackage{footmisc}
\usepackage{makecell}
\usepackage{bbding}
\usepackage{framed}
\usepackage{subfigure}
\usepackage{booktabs}
\usepackage{algorithm}
\usepackage{algorithmic}
\usepackage{tablefootnote}
\journal{Pattern Recognition}

\let\oldequation\equation
\let\oldendequation\endequation

\renewenvironment{equation}
{\linenomathNonumbers\oldequation}
{\oldendequation\endlinenomath}

\begin{document}

\begin{frontmatter}

\title{HierCode: A Lightweight Hierarchical Codebook for Zero-shot Chinese Text Recognition}

\author[mymainaddress]{Yuyi Zhang\corref{cor2}}
\ead{yuyi.zhang11@foxmail.com}

\author[secondaddress]{Yuanzhi Zhu\corref{cor2}}
\ead{z.yuanzhi@foxmail.com}

\author[mymainaddress]{Dezhi Peng}
\ead{pengdzscut@foxmail.com}

\author[mymainaddress]{Peirong Zhang}
\ead{eeprzhang@mail.scut.edu.cn}

\author[mymainaddress]{Zhenhua Yang}
\ead{eezhyang@gmail.com}

\author[secondaddress]{Zhibo Yang}
\ead{yangzhibo450@gmail.com}

\author[secondaddress]{Cong Yao}
\ead{yaocong2010@gmail.com}

\author[mymainaddress]{Lianwen Jin\corref{cor1}}

\ead{lianwen.jin@gmail.com}

\cortext[cor1]{Corresponding author.}
\cortext[cor2]{These authors contributed equally.}

\address[mymainaddress]{School of Electronic and Information Engineering, South China University of Technology, Guangzhou, China.}
\address[secondaddress]{Alibaba DAMO Academy, Hangzhou, China.}

\begin{abstract}
Text recognition, especially for complex scripts like Chinese, faces unique challenges due to its intricate character structures and vast vocabulary. 
Traditional one-hot encoding methods struggle with the representation of hierarchical radicals, recognition of Out-Of-Vocabulary (OOV) characters, and on-device deployment due to their computational intensity. 
To address these challenges, we propose HierCode, a novel and lightweight codebook that exploits the innate hierarchical nature of Chinese characters. 
HierCode employs a multi-hot encoding strategy, leveraging hierarchical binary tree encoding and prototype learning to create distinctive, informative representations for each character.
This approach not only facilitates zero-shot recognition of OOV characters by utilizing shared radicals and structures but also excels in line-level recognition tasks by computing similarity with visual features, a notable advantage over existing methods.
Extensive experiments across diverse benchmarks, including handwritten, scene, document, web, and ancient text, have showcased HierCode's superiority for both conventional and zero-shot Chinese character or text recognition, exhibiting state-of-the-art performance with significantly fewer parameters and fast inference speed. 

\end{abstract}





\end{frontmatter}


\section{Introduction}
\label{Introduction}
Text recognition is a fundamental task in computer vision that has been intensively studied for decades~\cite{shi2016end, xie2017pami, LIU2019Curved, chen2021scene, peng2022pagenet, Wang2023MM, YANG2024PR}.
Despite the success of prior methods of Chinese Text Recognition (CTR)~\cite{huang2021zero, Peng2022segment, lyu2022maskocr}, a majority of them still inherit the one-hot encoding paradigm from English text recognition approaches. 
This paradigm may not be the most suitable for CTR, primarily due to the following reasons:

First, one-hot encoding falls short in comprehensive feature representation of Chinese characters. 
Chinese script, known for its high information entropy among prevalent languages~\cite{wong1976comment}, holds significantly more information in its characters compared to Latin counterparts, especially in the intricate hierarchical information such as structures and radicals. 
Nevertheless, one-hot encoding allocates only one valid bit for each Chinese character, which fails to express the hierarchical richness within Chinese characters, leading to an extensive loss of critical structural and semantic information. 

Second, models reliant on one-hot encoding are incapable of zero-shot recognition, which is a crucial capability given the extensive and ever-growing lexicon of Chinese characters. 
For instance, the latest Chinese standard, GB18030-2022\footnote{https://openstd.samr.gov.cn/bzgk/gb/}, contains 87,887 categories, a significant increase from the 27,533 in the GB18030-2000 standard. 
Consequently, to fully recognize all Chinese characters, models are required to perform Out-Of-Vocabulary (OOV) character recognition, also known as zero-shot recognition, in which characters in testing are not seen in training. 
One-hot encoding, however, is inherently restricted to represent a limited range of characters, thus preventing models built based upon it from performing zero-shot recognition. 
To address the zero-shot problem, previous methods explored ways to leverage the glyph~\cite{HUANG2022Hipp, AO2022gl}, radicals~\cite{liu2013online, zhang2018radical, wang2019fewshotran, ZHANG2020radical, cao2020hde, LUO2023cue, LI2024side} or strokes~\cite{SU2003stroke, LIU20012stroke, chen2021stroke} information of Chinese characters.
However, these approaches primarily focus on character-level zero-shot recognition and do not easily extend to the recognition of text lines, a task associated with a broader spectrum of practical application scenarios.

Moreover, the one-hot encoding approach introduces significant barriers to deployment due to the expansive size of the classification layer required.
Generally, with an increasing number of character classes, the classification layer becomes excessively large, accounting for a disproportionate amount of the model's parameters. 
For example, in systems such as PP-OCR~\cite{du2020pp}, the classification layer for 20,000 character categories can constitute over 60\% of the model's total parameters. 
This presents considerable challenges in terms of computational efficiency and deployment on devices with limited resources. 
To decrease the parameter size, Hamming-OCR~\cite{li2020hamming} and EMU~\cite{li2022effective} propose different multi-hot encoding strategies as alternatives to one-hot encoding. 
However, these methods have not effectively captured the hierarchical information inherent in Chinese characters.

\begin{figure}[ht]
    \centering
    \begin{minipage}[b]{0.7\textwidth}
    \includegraphics[width=\textwidth]{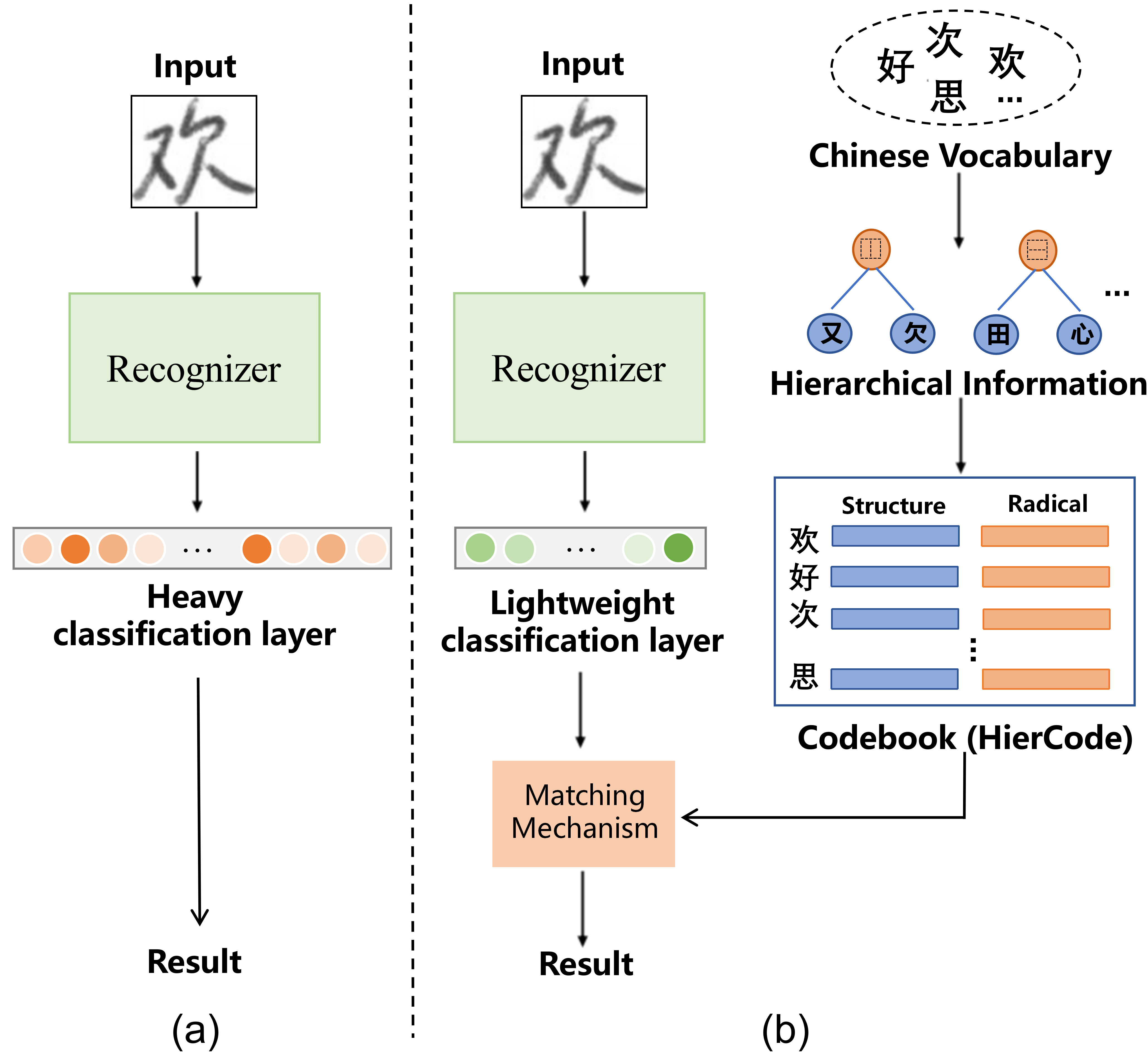}
    \end{minipage}
    \caption{Comparison between the framework of one-hot encoding methods (a) and that of the proposed method (b).}
    \label{Fig.intro}
\end{figure}

In this paper, we propose a novel hierarchical codebook \textbf{HierCode} to address all these issues. 
We first leverage the binary trees to represent hierarchical information within Chinese characters, including the structures and radicals, as shown in Fig.~\ref{Fig.intro}~(b). 
Then, we employed a RAN~\cite{wang2019fewshotran} model to derive a full set of radical prototypes, from which all Chinese characters can be composed. 
Subsequently, we generate a unique and robust multi-hot encoding representation for each Chinese character. 
By culminating the encodings of all characters, we establish an integrated codebook named HierCode.
In the training phase, HierCode is employed to supervise the recognition model, which is a traditional encoder-decoder framework but substitutes the one-hot classification layer with multi-hot alternatives. 
In the inference phase, HierCode's multi-hot encoding allows the model to compute the similarity between HierCode and visual features to match the ultimate prediction for each character in a textual image. 
This mechanism inherently supports character-level zero-shot recognition and can be seamlessly applied to line-level recognition tasks.
Simultaneously, the multi-hot approach also ensures that HierCode remains lightweight, using fewer bits to represent characters, thus increasing the model's inference speed without compromising on performance.
Through extensive experiments on a variety of benchmarks, including handwritten, scene, document, web, and ancient texts, HierCode has demonstrated superior performance in both standard and zero-shot CTR tasks. The results reveal that HierCode not only outperforms many existing methods but also offers advantages in terms of model footprint and inference efficiency. 

The contributions of this paper are three-fold:
\begin{itemize}
\setlength{\itemsep}{0pt}
    \item We propose a hierarchical codebook named HierCode, which provides unique and informative representations for each Chinese character through hierarchical encoding and prototype learning. 
    \item The hierarchical combination of radical features of Chinese characters enables the model to deal with the zero-shot Chinese recognition at both character and line levels. 
    Moreover, the multi-hot encoding employed by HierCode exhibits lightweight characteristics and fast inference speed, thereby significantly enhancing its practical applicability.
    \item Extensive experiments conducted on diverse datasets demonstrated that HierCode not only achieves state-of-the-art accuracy in zero-shot Chinese character recognition and outperforms the majority of existing approaches in Chinese text recognition, but also exhibits fast inference speed and small footprint, facilitating the development of lightweight Chinese text recognition networks. 
\end{itemize}

\section{Related Work}
\label{Related Work}

\subsection{Chinese Character Recognition Methods}
Chinese Character Recognition (CCR) methods have experienced significant development over several decades.
Early CCR methods mainly relied on hand-crafted features~\cite{jin2001study, su2003novel, chang2006techniques}.
With the advancements in deep learning, Convolutional Neural Networks (CNN) are now widely used for feature extraction and achieve exceptional performance~\cite{cirecsan2015multi, xiao2017building, LI2020gl}. 
Although CNN-based techniques generally prove effective, they tend to struggle with zero/few-shot CCR problems.
Given that Chinese characters can be hierarchically decomposed into sequences of structures and radicals, numerous zero/few-shot recognition techniques transform the CCR problem into a sequence prediction problem, thereby enabling the recognition of Out-Of-Vocabulary (OOV) characters.
For instance, Wang et al.~\cite{wang2017radical} utilized multi-label learning to recognize radicals.
Zhang et al.~\cite{zhang2018radical} and Wang et al.~\cite{wang2018denseran} employed spatial attention mechanisms to decode structures and radicals sequences from image features. 
Wang et al.~\cite{wang2019fewshotran} mapped radicals to the feature space and aggregated radical features via prototype learning. 
Cao et al.~\cite{cao2020hde} considered the hierarchical decomposition information while designing embedding rules for Chinese characters, facilitating zero-shot CCR. 
Chen et al.~\cite{chen2021stroke} decomposed Chinese characters into finer strokes and utilized printed character image matching to tackle the one-to-many problem between stroke sequences and Chinese characters.
Luo et al.~\cite{LUO2023cue} quantified the significance of radicals in CCR from the information theory perspective, thereby improving recognition accuracy in zero-shot CCR experiments.
While these methods effectively address zero/few-shot recognition for Chinese characters, their complex networks and decoding strategies render them unsuitable for line-level recognition, essentially damaging their practical applicability.

\subsection{Chinese Text Recognition Methods}
Initially, Chinese Text Recognition (CTR) methods output entire text line results through character recognition based on sliding window~\cite{wang2012end} and segmentation techniques~\cite{bissacco2013photoocr, jaderberg2014deep}. 
Subsequently, Shi et al.~\cite{shi2016end} introduced CRNN, treating the text line holistically and predicting character sequences directly from the input image via an encoder-decoder framework.
Later, various methods~\cite{hu2020gtc,yousef2020accurate,chen2020multrenets} enhanced and extended the CRNN framework, applying it to CTR~\cite{xie2017pami, liu2021searching, Z_Wang_Writer}.
Concurrently, attention-based methods emerged~\cite{shi2018aster,luo2019moran,wang2020decoupled, LIN2021STAN}, achieving breakthroughs in irregular text recognition. 
Meanwhile, there are some new methods~\cite{fang2021read,wang2022petr,Peng2022segment} incorporate powerful Transformer-based language models~\cite{vaswani2017attention} to improve performance.
However, despite their commendable performance in Latin benchmarks, these methods encounter significant challenges when applied to Chinese benchmarks.
Specifically, these methods primarily depend on the one-hot encoding strategy, which falls short of capturing the hierarchical information embedded within Chinese characters. 
Moreover, these techniques do not excel in zero-shot recognition. 
To tackle these challenges, ZCTRN~\cite{huang2021zero} introduced a class embedding module inspired by HDE~\cite{cao2020hde}, and Yu et al.~\cite{yu2023ctrclip} devised a pre-trained CLIP-like model for aligning printed character images and ideographic description sequences. 
These methods generate canonical feature representations for each Chinese character and achieve zero-shot recognition at the line level by matching visual features with character representations. 
However, their complex architectures lead to a significant increase in the parameter size for these methods. 
This inspires us to explore a more lightweight encoding method for Chinese characters.

\subsection{Lightweight Text Recognition Methods}
In recent times, there has been extensive research on developing lightweight text recognition methods to enable the deployment of text recognition algorithms on mobile devices. 
For example, PP-OCR~\cite{du2020pp} reduced the model's weight by decreasing the number of channels in CRNN~\cite{shi2016end}.
Hamming-OCR~\cite{li2020hamming} introduced a hash encoding approach for Chinese characters, replacing the traditional one-hot encoding with multi-hot encoding. 
This approach significantly reduced the parameters of the classification layer in text recognizers. 
Consequently, it achieved a smaller model size compared to PP-OCR. 
Based on Hamming-OCR, EMU~\cite{li2022effective} implemented hash encoding for Chinese characters and employed a progressive binarization strategy to improve recognition accuracy compared to Hamming-OCR. 
While these previous methods effectively reduced the model's parameters, they often resulted in decreased performance.


\section{Methodology}
\label{Methodology}

\subsection{Hierarchical Representation of Chinese Characters}
\label{Sec_Hierarchical}

\begin{figure}[ht]
    \centering
    \begin{minipage}[b]{0.95\textwidth}
    \includegraphics[width=\textwidth]{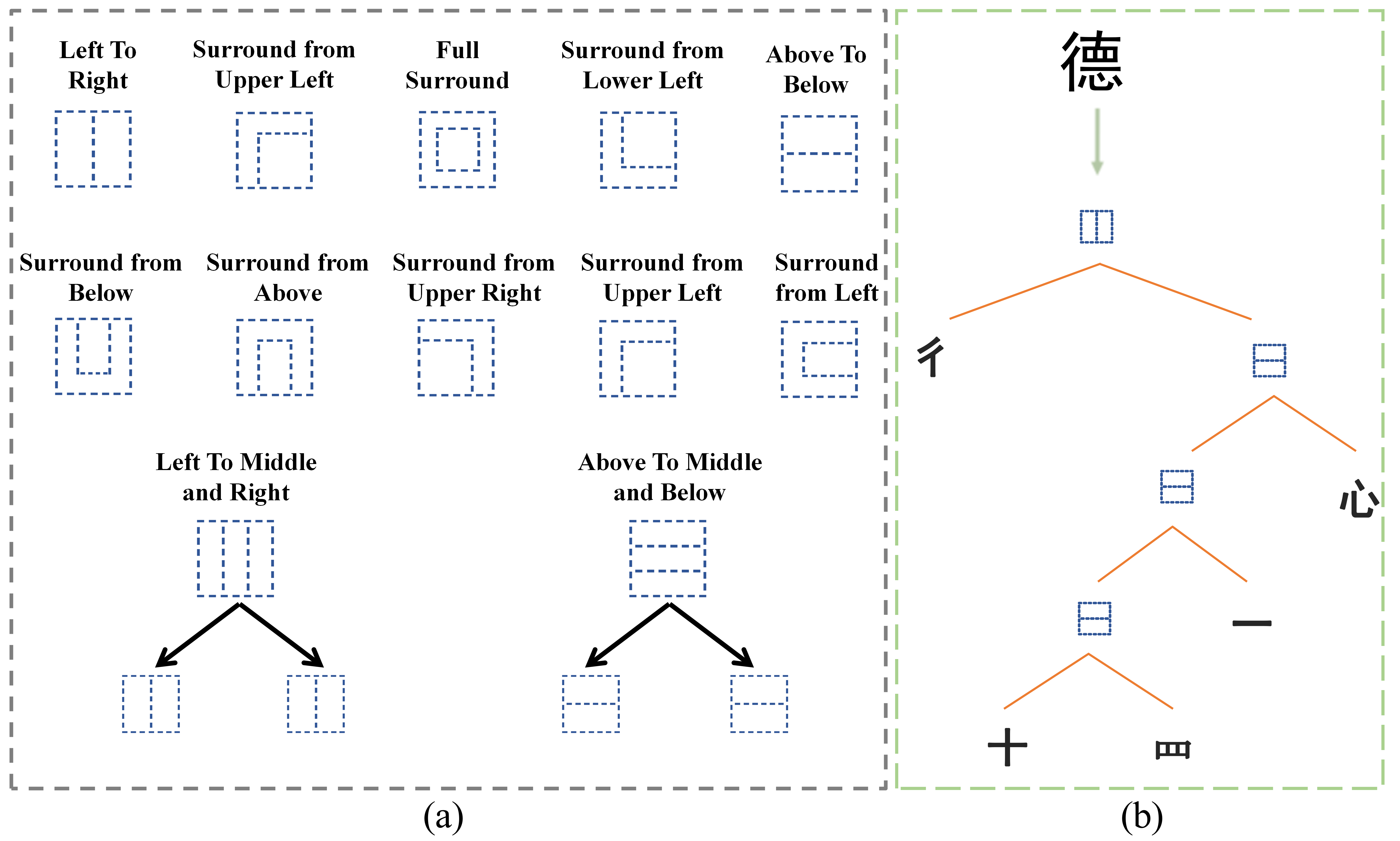}
    \end{minipage}
    \vspace{-2mm}
    \caption{Illustration of (a) 12 Chinese character structures and (b) the hierarchical decomposition of a Chinese character.}
    \label{Fig.2}
\end{figure}

In contrast to English and Latin language, Chinese characters embody a more complex structure, each comprising a set of radicals characterized by unique spatial arrangements~\cite{zhang2018radical, wang2018denseran, wang2019fewshotran}.
These radicals and their configurations can be systematically categorized into twelve distinct structural types as per Unicode standards, such as above-to-below alignment, left-to-right alignment, etc., as shown in Fig.~\ref{Fig.2}~(a). 
For practicality and computational efficiency, our methodology simplifies the encoding of certain ternary structures by deconstructing them into binary equivalents, ultimately consolidating them into ten primary structural categories for our encoding purposes.

To encode each character, we first decompose it into discrete structures and radicals. 
Then we unfold these components with a binary tree, where structures reside at non-leaf nodes and radicals reside at leaf nodes, as shown in Fig.~\ref{Fig.2}~(b). 
This results in unique binary trees for different characters, varying in width, depth, and node arrangements. 
Given that the set of radicals and structures is universally shared among all Chinese characters, this hierarchical representation is comprehensive, capturing the entirety of the Chinese lexicon.

\subsection{The Construction of HierCode}
\label{Sec_METH_2}

\begin{figure*}
\centering
\includegraphics[width=1\textwidth]{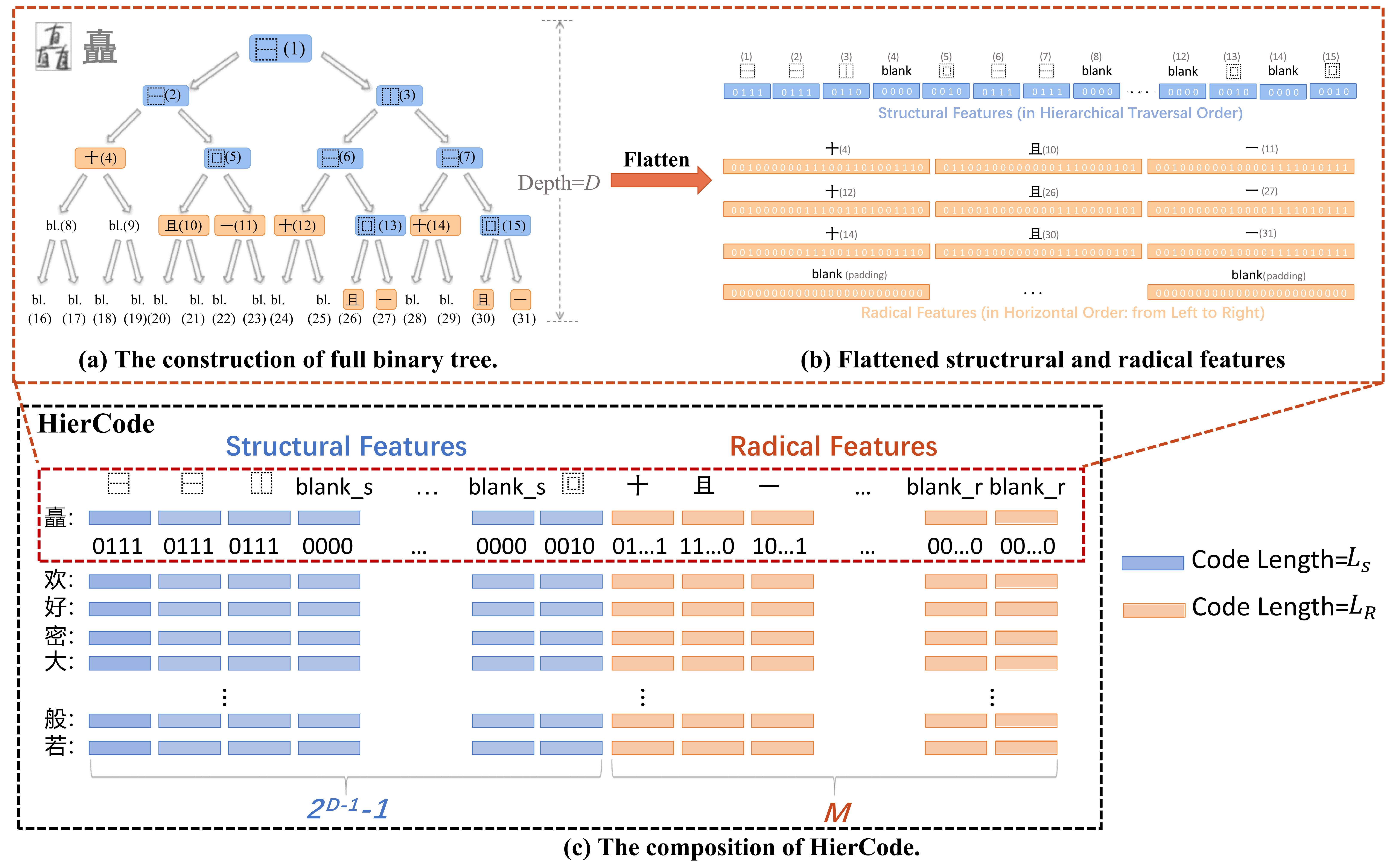}
\vspace{-10mm}
\caption{A schematic overview of HierCode. The `bl.' in (a) represents the empty node used to fill the binary tree.}
\label{Fig.3}
\vspace{-3mm}
\end{figure*}

We illustrate the overview of the construction process of the proposed HierCode in Fig.~\ref{Fig.3}, which is a multi-hot codebook consisting of the hierarchical representations of a given set of Chinese characters. 

First, according to Sec.~\ref{Sec_Hierarchical}, each Chinese character has a unique hierarchical binary tree with distinctive width and depth. 
To accommodate different characters into a codebook, we normalize the character encoding lengths by setting a blank binary tree with a maximum depth of $D$ and a maximum width of $2^{D-1}$, and denote it as the full tree. 
We then extract the hierarchical representation for each character and populate the full tree accordingly, as depicted in Fig.~\ref{Fig.3}~(a). 
Second, corresponding to the structures and radicals in the hierarchical features, we extract the \textit{structural features} and \textit{radical features} for each Chinese character.



\textbf{Structural Features.} 
The structural features capture the structural information of the Chinese character. 
We establish encodings of fixed length $L_\mathbf{S}$ for the 10 structures outlined in Sec.~\ref{Sec_Hierarchical}, resulting in the formation of a structural code set \textbf{S}. 
Given the constraint of a fixed number of character structures (set to 10), we can efficiently encode them using just four bits. 
Therefore, we set $L_\mathbf{S}$ as 4 and manually assign codes to each structure. 
Subsequently, to obtain the encoding for the structural features code, we traverse the nodes of the binary tree by the breadth-first algorithm, as shown in Fig.~\ref{Fig.3}(a). 
During this traversal, if a structure is encountered on a node, the node is represented by its corresponding structure code from the structural code set \textbf{S}. 
On the other hand, if a radical or an empty node is encountered, the node is represented by an all-zero code $blank_\mathbf{S}$ with length $L_\mathbf{S}$. 
The final result is the encoding of the structural features, depicted in the blue region of Fig.~\ref{Fig.3}(b).
Since structures can only appear on non-leaf nodes, given the maximum depth $D$, there are a total of $2^{D-1}-1$ structure codes.
Therefore, the structural features $\mathbf{C_S}$ can be formulated as:
\begin{equation}
\mathbf{C_S} = [S_1, S_2, ... , S_{2^{D-1}-1}], S_i \in \mathbf{S} \cup blank_\mathbf{S}.  
\end{equation}

\textbf{Radical Features.} 
We leverage prototype learning to acquire unique and informative representations for radicals, in which a RAN~\cite{wang2019fewshotran} is trained to extract radical prototypes.
Subsequently, each radical prototype is binarized into -1 or 1, generating the set of radical codes $\textbf{R}$.
We extract the radicals from the binary tree in the hierarchical traversal order, convert them to the corresponding code from set $\textbf{R}$, and arrange them horizontally from left to right, as shown in Fig.~\ref{Fig.3}(b).
Furthermore, considering that various Chinese characters consist of differing numbers of radicals, we pad the all-zero code $blank_\mathbf{R}$ with length $L_\mathbf{R}$ to the maximum number of radicals $M$.
Therefore, the radical features $\mathbf{C_R}$ can be represented as:
\begin{equation}
\mathbf{C_R} = [R_1, R_2, ... , R_M], R_i \in \mathbf{R} \cup blank_\mathbf{R}.  
\end{equation}

The complete multi-hot encoding $\mathbf{C}$ for each character consists of structural features $\mathbf{C_S}$ and radical feature $\mathbf{C_R}$, and can be described as:

\begin{equation}
\mathbf{C} = (c_1, c_2, ... , c_t) = (\mathbf{C_S}, \mathbf{C_R}), c_i \in \{-1, 1\}, 
\end{equation}
where $(\cdot)$ represents the concatenation process for different code bits or sequences. 
$i$ is index and $c_i$ denotes the $i$-th bit in $\textbf{C}$.
$t$ is the length of multi-hot encoding and can be calculated as:
\begin{equation}
\label{eq_t}
t = (2^{D-1}-1) \times L_\mathbf{S} + M \times L_\mathbf{R}. 
\end{equation}
where $L_\mathbf{S}$ and $L_\mathbf{R}$ represent the length of structural and radical code, respectively.
$M$ represents the maximum number of radicals in the Chinese character with the most radicals.
By culminating the multi-hot encodings of all characters, we establish the HierCode $\mathbf{H}$ which can be expressed as:
\begin{equation}
\mathbf{H} = \begin{bmatrix}C_1, C_2, \dots, C_N
\end{bmatrix}^T
\end{equation}
where $N$ represents the number of character categories. Refer to Fig.~\ref{Fig.3}~(c) for a more lucid visual depiction.

\subsection{Text Recognition with HierCode}

\begin{figure*}
\centering
\vspace{-3mm}
\includegraphics[width=0.95\textwidth]{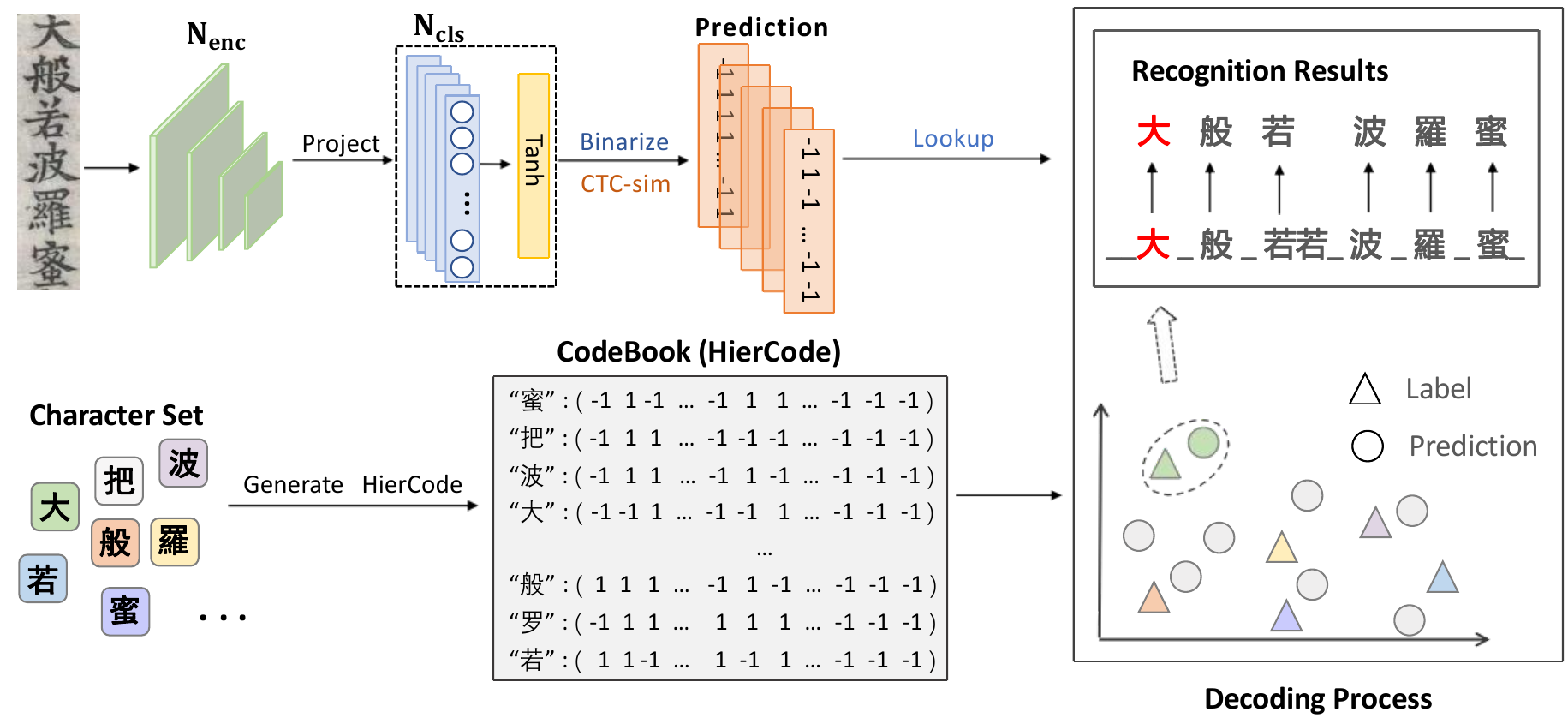}
\caption{The overall architecture for text recognition using HierCode.}
\label{Fig.5}
\vspace{-4mm}
\end{figure*}

Fig.~\ref{Fig.5} illustrates the pipeline for text recognition using HierCode.
Given a text image $x$ as input and outputs a sequence of visual features $\mathbf{V} = \{v_1, v_2, ... , v_W\}$, with $v_i \in \mathbb{R}^C$, where $W$ represent the width of the feature map and $C$ is the output channel size. 
Subsequently, the multi-hot classification layer $N_{cls}$ performs a binarization operation on $\mathbf{V}$ and outputs a sequence of binarized vectors, which can be expressed as:

\begin{equation}
\mathbf{B} = \{b_1, b_2, ... , b_W\} = N_{cls}(\mathbf{V}), b_i \in \{-1, 1\}^C
\end{equation}

Specifically, the multi-hot classification layer consists of a fully connected layer and an activation function. 
The activation function can be Sigmoid or Tanh, which activate each bit independently. 
Our empirical study shows that the two activation functions perform comparably, and in this work, we use Tanh. 

To solve the alignment problem caused by unequal label lengths of text, we modify the vanilla CTC loss~\cite{graves2009novel} and propose a similarity-based CTC loss $L_{CTC-sim}$ to train the HierCode-based text recognition model, which can be expressed as:

\begin{equation}
\label{eq_sim}
d(\mathbf{H}^T, \mathbf{B})=\mathbf{H}^T \cdot \mathbf{B},  \mathbf{H} \in \mathbb{R}^{C\times N}, \mathbf{B} \in \mathbb{R}^{C\times W}
\end{equation}

\begin{equation}
L_{CTC-sim}=-\sum \log p\left(l \mid d(\mathbf{H}^T, \mathbf{B})\right),
\end{equation}

\noindent where $l$ is the ground truth label, and $d(\mathbf{H}^T, \mathbf{B})$ is the inner product to measure the similarity between HierCode $\mathbf{H}$ and predicted multi-hot code $\mathbf{B}$. 

When HierCode is employed in character-level, attention-based, and Transformer-based text recognizers, we propose a similarity-based cross-entropy loss $L_{CE-sim}$ as: 
\begin{equation}
L_{CE-sim}=-\sum \log p\left(l \mid d(\mathbf{H}^T, \mathbf{B})\right),
\end{equation}

In the inference stage, we obtain the predicted codes and calculate their similarity with each character in the codebook by Eq.~\ref{eq_sim}. 
The character exhibiting the highest similarity is selected as the final prediction.


\section{Experiments}
\subsection{Datasets}

\textbf{ICDAR2013}~\cite{yin2013icdar} is a handwritten Chinese competition database, which contains subsets of text line data (denoted as ICDAR-line) and isolated character data (denoted as ICDAR-char), and we use these two subsets as the test set.

\textbf{CASIA-HWDB}~\cite{liu2011casia} is a large-scale Chinese handwritten database. In this study, we use the text line part~(HWDB 2.0-2.2) and the isolated character part~(HWDB 1.0-1.2) as the training set for ICDAR2013. 

\textbf{BCTR}~\cite{chen2021bctr} is a large-scale Chinese text image benchmark, which consists of four subsets, i.e., scene, web, document~(denoted as Doc), and handwriting~(denoted as Handw). 

\textbf{MTHv2}~\cite{ma2020joint} contains contains 105,579 text line images collected from ancient Chinese scriptures. 

\textbf{CTW}~\cite{Yuan2019ctw} contains 812,872 Chinese character instances collected from street views.

\subsection{Implementation Details} 
\textbf{Network Architecture.}
For character-level recognition, we adopt ResNet18~\cite{he2016deep} as our backbone.
For line-level recognition, we utilize ResNet34~\cite{he2016deep} as our backbone, followed by a BLSTM~\cite{Hochreiter1997lstm} layer.
The proposed hyper-parameters $D$, $L_\mathbf{S}$, $L_\mathbf{R}$, $M$, and $t$ are set to 5, 4, 36, 9, and 384, respectively, and will be analyzed in Sec.~\ref{Sec_Ablation}. 
In text line recognition experiments, the height of the training images is resized to 128, and the width is calculated with the original aspect ratio (up to 1920). 
The data pre-processing strategy described in~\cite{Peng2022segment} is adopted in the ICDAR-line. Specifically, we rectify the text-line image to make the text horizontal and remove white padding on the top and bottom of the image to highlight the text.
For the character recognition, each input image is resized to 96$\times$96. 

\textbf{Optimization.}
The proposed method is implemented using PyTorch. 
We apply the Adadelta optimizer with an initial learning rate of 0.1.
The batch size is set to 128. 
All experiments are conducted on NVIDIA RTX 1080Ti GPU with 11 GB memory.

\begin{table}[t]
\small
\centering
\vspace{-3mm}
\caption{Comparison of recognition performance~(\%), speed, and model size with previous methods on the ICDAR-line dataset. }
\vspace{1mm}
\begin{tabular}{lrrrrrr}
\hline
{\textbf{Methods}} & \textbf{AR ↑} & \textbf{CR ↑} & \textbf{Speed ↑} & \textbf{Model Size ↓}      \\
\hline
Messina et al.~\cite{R_Messina_Segmentation} & 83.50 & - & - & - \\
Wu et al.~\cite{Y_Wu_Handwritten}   & 86.64 & 87.43 & - & \underline{71MB} \\
Du et al.~\cite{J_Du_Deep}    & 83.89 & - & - & - \\
Wang et al.~\cite{S_Wang_Deep} & 88.79 & 90.67 & - & - \\
Wang et al.~\cite{Z_Wang_A_Comprehensive} & 89.66 & - & - & - \\
Xie et al.~\cite{Z_Xie_Aggregation}   & 91.25 & 91.68 & - & - \\
Peng et al.~\cite{D_Peng_Fast}  & 89.61 & 90.52 & - & - \\
Xiu et al.~\cite{Y_Xiu_A_Handwritten} & 88.74 & - & - & -  \\
Xie et al.~\cite{C_Xie_High} & 91.55 & 92.13 & 64fps  & \textbf{61MB}  \\
Wang et al.~\cite{Z_Wang_Writer} & 91.58 & - & - & -  \\
Wang et al.~\cite{Z_Wang_Weakly} & 87.00    & 89.12 & - & -  \\
Zhu et al.~\cite{Z_Zhu_Attention} & 90.86 & - & - & -   \\
Liu et al.~\cite{huang2021zero}  & 93.62 & - & - & 203MB \\
Peng et al.~\cite{Peng2022segment}  & \underline{94.50} & \underline{94.76} & \underline{70fps} & 119MB \\
HierCode (Ours) & \textbf{94.68} & \textbf{95.11} & \textbf{161fps} & 90MB \\
\hline
\label{Table_ICDAR13}
\end{tabular}%
\vspace{-8mm}
\end{table}%

\textbf{Evaluation.}
For experiments on the CASIA-HWDB and ICDAR-line dataset, the evaluation criteria are accuracy rate (AR) and correct rate (CR) specified by ICDAR2013 competition~\cite{yin2013icdar}. 
For experiments on the BTCR dataset, we follow the same process as in~\cite{lyu2022maskocr} and compute the accuracy in sentence level over each subset and the whole dataset. 
We further evaluate the recall of Chinese characters~(ReC) and non-Chinese characters~(ReL) in the text line experiments. 
Non-Chinese characters mainly include Latin, numbers and symbols, etc., and their performance is used to verify the generalization of HierCode in text recognition. 
Furthermore, AR-zero refers to the evaluation criteria of line-level zero-shot setting in ZCTRN~\cite{huang2021zero}. 
For character recognition related experiments, we use character accuracy as the evaluation criteria.

\subsection{Recognition Performance Comparison with State-of-the-Art Methods}

\textbf{Handwritten Chinese text recognition.} 
We commence with experiments on the ICDAR-line dataset, with results illustrated in Tab.~\ref{Table_ICDAR13}. Compared with the SOTA method~\cite{Peng2022segment}, our method improves AR by 0.18\% and CR by 0.35\%, while also achieving the highest inference speeds and smaller model sizes significantly.

\begin{table}[t]
\setlength\tabcolsep{3pt}
\small
\centering
\vspace{-1mm}
\caption{Comparison of recognition accuracy in sentence level~(\%) with previous methods on the BCTR dataset. `*' indicates that the method uses additional data. The numbers of `$\Delta$' in \textcolor[RGB]{0,175,79}{\textbf{green}} and \textcolor{blue}{\textbf{blue}} denote the improvements over each subset and average~(Avg), respectively.
The first eight results are derived from~\cite{lyu2022maskocr}.
}
\vspace{1mm}
\begin{tabular}{lrrrrr}
\hline
 \textbf{Methods} & \textbf{Scene} & \textbf{Web}   & \textbf{Doc} & \textbf{Handw} & \textbf{Avg} \\
\hline
 CRNN~\cite{shi2016end}  & 53.4  & 54.5  & 97.5  & 46.4  & 67.0     \\
 ASTER~\cite{shi2018aster} & 54.5  & 52.3  & 93.1  & 38.9  & 64.7   \\
 MORAN~\cite{luo2019moran}  & 51.8  & 49.9  & 95.8  & 39.7  & 64.3   \\
 SAR~\cite{Li2019show}    & 62.5  & 54.3  & 93.8  & 31.4  & 67.3   \\
 SRN~\cite{yu2020towards}    & 60.1  & 52.3  & 96.7  & 18.0    & 65.0     \\
 SEED~\cite{qiao2020seed}   & 49.6  & 46.3  & 93.7  & 32.1  & 61.2   \\
 TransOCR~\cite{chen2021scene}  & 63.3  & 62.3  & 96.9  & 53.4  & 72.8   \\
 MaskOCR*~\cite{lyu2022maskocr} & \textit{76.2} & \textit{76.8} & \textit{99.4} & \textit{67.9} & \textit{82.6}  \\
 ABINet~\cite{fang2021read}  & \textbf{64.4}  & \textbf{67.4}  & \underline{97.2}  & \underline{54.8}  & \underline{74.1}   \\

\hline
 One-hot (Baseline) & 60.3      & 60.2      & 92.8      & 54.1      & 70.0        \\
 HierCode (Ours)  & \underline{63.7}  & \underline{66.2}  & \textbf{98.2}  & \textbf{56.3}  & \textbf{74.2}  \\
$\Delta$ & \textcolor[RGB]{0,175,79}{\textbf{+3.4}} & \textcolor[RGB]{0,175,79}{\textbf{+6.0}} & \textcolor[RGB]{0,175,79}{\textbf{+5.4}} & \textcolor[RGB]{0,175,79}{\textbf{+2.2}} & \textcolor{blue}{\textbf{+4.2}}  \\        
\hline     
\label{Table_BCTR}
\end{tabular}%
\vspace{-10mm}
\end{table}%

\textbf{Multi-scenario text recognition.} 
We further evaluate the efficacy of HierCode across a broader spectrum of text recognition scenarios, which comprises four distinct text types: scene, web, document, and handwritten. Results are given in Tab.~\ref{Table_BCTR}. 
Compared to the one-hot encoding baseline, HierCode demonstrated notable improvements in recognition accuracy across all scenarios: an increase of 3.4\% for scene text, 6.0\% for web text, 5.4\% for document text, and 2.2\% for handwritten text. 
These results underscore the versatility and robustness of HierCode in handling diverse text recognition tasks. 

When compared against prevailing methods, our proposed HierCode sets new records on the document and handwriting datasets. 
While in the realm of scene and web text recognition, HierCode's performance was marginally outpaced by the state-of-the-art ABINet model~\cite{fang2021read}. 
This minor discrepancy in performance may be attributed to the complexities and diverse backgrounds present in scene and web data, which renders larger recognition difficulty for our relatively simpler backbone. 
In contrast, ABINet is specifically tailored for scene text recognition, reasonably resulting in better performances. 
It should be noticed that although MaskOCR~\cite{lyu2022maskocr} demonstrates enhanced performance metrics, its reliance on an extensive corpus of additional data for pretraining introduces an unfairness in comparison, given that our HierCode approach does not engage in a pretraining phase.

\begin{table}[t]
\setlength\tabcolsep{5pt}
\small
\centering
\vspace{-2mm}
\caption{Comparison of the recall rate~(\%) of Chinese~(ReC) and Non-Chinese characters~(ReL) between one-hot and HierCode. The numbers of `$\Delta$' denote the improvements over each subset.}
\vspace{1mm}
\begin{tabular}{llrrrrr}
\hline
\multirow{2}[0]{*}{\textbf{Metric}} & \multirow{2}[0]{*}{\textbf{Methods}} & \multirow{2}[0]{*}{\textbf{ICDAR-line}} & \multicolumn{4}{c}{\textbf{BCTR}} \\
\cline{4-7}          &       &       & \textbf{Scene} & \textbf{Web}   & \textbf{Doc} & \textbf{Handw}\\
\hline
\multirow{3}[0]{*}{\textbf{ReC}} & One-hot & 93.35 & 82.09 & 79.57 & 98.64      & 91.65 \\
& HierCode & 94.53 & 83.41 & 83.39 & 99.71  & 92.35 \\
& $\Delta$ & \textcolor[RGB]{0,175,79}{\textbf{+1.18}}  & \textcolor[RGB]{0,175,79}{\textbf{+1.32}}  & \textcolor[RGB]{0,175,79}{\textbf{+3.82}}  & \textcolor[RGB]{0,175,79}{\textbf{+1.07}}      & \textcolor[RGB]{0,175,79}{\textbf{+0.70}}   \\
\hline
\multirow{2}[0]{*}{\textbf{ReL}} & One-hot & 85.56 & 90.24 & 84.67 & 99.37      & 86.59  \\
& HierCode & 85.65 & 90.27 & 85.19 & 99.54  & 86.61  \\
& $\Delta$ & \textcolor[RGB]{0,175,79}{\textbf{+0.09}}  & \textcolor[RGB]{0,175,79}{\textbf{+0.03}}  & \textcolor[RGB]{0,175,79}{\textbf{+0.52}}  & \textcolor[RGB]{0,175,79}{\textbf{+0.17}}      & \textcolor[RGB]{0,175,79}{\textbf{+0.02}}   \\
\hline
\end{tabular}%
\label{Tab_per_detail}%
\vspace{-2mm}
\end{table}%

Furthermore, we analyze the line-level recognition performance changes in Chinese and non-Chinese languages as presented in Tab.~\ref{Tab_per_detail}. 
In each dataset, we observe that the performance of Latin characters, numbers, and symbols on HierCode is on par with that of one-hot methods, whereas HierCode exhibits a significant improvement in the recognition of Chinese characters. 
This observation substantiates the premise that HierCode's performance benefits stem primarily from its hierarchical representation of Chinese characters, which is in concordance with the foundational design objectives of our method. 


\subsection{Zero-shot Capability}
\label{Sec_Zero}

\textbf{Zero-shot character-level recognition.} 
We keep the same settings as the previous method~\cite{wang2018denseran,cao2020hde,chen2021stroke, LUO2023cue}. 
Specifically, for the handwritten characters, we use HWDB1.0-1.1 and ICDAR-char, which consist of 3,755 classes. 
We select first $m$ classes from HWDB1.0-1.1 as the training set, where $m$ ranges in \{500, 1000, 1500, 2000, 2755\}. 
The test set is composed of samples from the last 1000 classes of ICDAR-char.
For the scene characters, we use CTW dataset and choose samples of the first $m$ classes as the training set, where $m$ ranges in \{500, 1000, 1500, 2000, 3150\} and choose samples of the last 1000 classes as the test set. 
In the training phase, we calculate the $L_{CE-sim}$ only for the characters that appear in the training set. 
In the inference phase, the final classification results are obtained by matching the model predictions with the full set of characters comprising both the training and test sets. 
In addition, methods~\cite{liu2023towards, liu2022open, yu2023ctrclip} using additional glyph support samples during the training or pre-training process are outside the scope of this study.

As is shown in Tab.~\ref{Table 1}, on the handwritten dataset, our method shows significant improvement compared to the SOTA method SideNet~\cite{LI2024side}, with absolute accuracy gains of 1.12\%, 4.51\%, 1.59\%, 1.57\% and 5.91\% at $m$ in \{500, 1000, 1500, 2000, 2755\}, respectively. On the scene dataset, HierCode achieves SOTA performance. 
Furthermore, we observe that even on the full class evaluation on the handwritten dataset, HierCode can still improve the performance by 0.54\% compared to the advanced method~\cite{cao2020hde} while maintaining the smallest model size.

\begin{table*}[t]
\small
\centering
\caption{Comparison of the character zero-shot setting on the handwritten dataset ICDAR-char and scene dataset CTW with the previous method. 
}
\vspace{1mm}
\resizebox{1\textwidth}{!}{%
\begin{tabular}{lrrrrrrrrrrrrr}
\hline
& \multicolumn{5}{c}{\textbf{Handwritten/\%~($m$ for the classes)}}               &       & \multicolumn{5}{c}{\textbf{Scene/\%~($m$ for the classes)}}                 & {\textbf{Full Class}} & \textbf{Model} \\
\cline{2-6}\cline{8-12}          & 500   & 1000  & 1500  & 2000  & 2755  &       & 500   & 1000  & 1500  & 2000  & 3150  & \textbf{Accuracy/\% }      & \textbf{Size} \\
\hline
DenseRAN~\cite{wang2018denseran}  & 1.70   & 8.44  & 14.71 & 19.51 & 30.68 &       & 0.15  & 0.54  & 1.60   & 1.95  & 5.39  &  96.66     & 287.9MB \\
HDE~\cite{cao2020hde}   & 4.90   & 12.77 & 19.25 & 25.13 & 33.49 &       & 0.82  & 2.11  & 3.11  & 6.96  & 7.75  &  \underline{97.14}     &  - \\
SLD~\cite{chen2021stroke}   & 5.60   & 13.85 & 22.88 & 25.73 & 37.91 &       & 1.54  & 2.54  & 4.32  & 6.82  & 8.61  &  96.73     &  \underline{287.4MB} \\
CUE~\cite{LUO2023cue} & \textbf{7.43}   & 15.75 & 24.01 & 27.04 & 40.55 &       & -  & -  & -  & -  & -  &  96.96     &  - \\
SideNet~\cite{LI2024side} & 5.10   & \underline{16.20} & \underline{33.80} & \underline{44.10} & \underline{50.30} &       & -  & -  & -  & -  & -  &  -    &  - \\
HierCode (Ours)  & \underline{6.22}  & \textbf{20.71} & \textbf{35.39} & \textbf{45.67} & \textbf{56.21} &       & \textbf{1.67}  & \textbf{2.59}  & \textbf{4.54}  & \textbf{7.02}  & \textbf{9.13}  &  \textbf{97.68}     &  \textbf{44.2MB} \\
\hline
\end{tabular}%
}
\label{Table 1}
\end{table*}%

\textbf{Zero-shot line-level recognition.} 
The ancient text dataset contains numerous rare and OOV characters, which can effectively verify the capability of HierCode in learning character structures and radicals. 
We conduct experiments on MTHv2 and follow the official training protocol. 
The results presented in Tab.~\ref{Table 3} indicate that our method achieves the best recognition performance with the smallest model size. 
In addition, benefiting from the hierarchical representation of characters, HierCode has zero-shot recognition ability, i.e., recognizing OOV characters, which is not available in most existing SOTA methods~\cite{shi2016end, ma2020joint}. 
We follow the zero-shot setting in ZCTRN~\cite{huang2021zero} that OOV characters are present in the text line data, allowing us to assess the zero-shot text line recognition capability of our method. 
Compared with ZCTRN, our method exhibits a significant performance improvement of 2.65\%. 
These results underscore the superior zero-shot recognition ability of the proposed HierCode. 

\begin{table}[t]
\small
\vspace{-4mm}
\caption{Comparison of recognition performance~(\%) and model size with previous methods on ancient text dataset MTHv2.}
\vspace{1mm}
\centering
\begin{tabular}{lrrrr}
\hline
\textbf{Methods} & \textbf{AR} & \textbf{CR} & \textbf{AR-zero} & \textbf{Model Size} \\
\hline
CRNN~\cite{shi2016end}  & 96.94 &97.15 & - &134.7MB  \\
RAN~\cite{zhang2018radical}    & 91.56    & 91.79  & 37.22 &\underline{107.6MB} \\
Ma et al.~\cite{ma2020joint}  & 95.52 &96.07 & - &- \\
ZCTRN~\cite{huang2021zero}  & \underline{97.42} &\underline{97.62} & \underline{51.40} &129.7MB  \\
HierCode (Ours)    & \textbf{97.87} &\textbf{98.05} & \textbf{54.05} &\textbf{56.8}MB \\
\hline
\label{Table 3}
\vspace{-4mm}
\end{tabular}
\end{table}


\subsection{Lightweight Characteristics}

The proposed HierCode has a significant advantage in model size compared to previous methods, as shown in Tab.~\ref{Table_ICDAR13}, Tab.~\ref{Table 3} and Tab.~\ref{Table 1}. 
This is due to the multi-hot encoding method employed in HierCode, which effectively reduces the number of parameters in the recognizer classification layer. 
In this subsection, we conduct experiments on HWDB1.0-1.1 and ICDAR-char dataset to investigate the compression capability of HierCode on lightweight backbones. 
Concretely, we integrate our method with two lightweight backbones, i.e., MobileNet v3 large and MobileNet v3 small, to evaluate the feasibility of our approach on mobile devices. 
The results in the `P-Cls' row of Tab.~\ref{Table 6}~show that applying HierCode can compress the number of parameters by approximately 92.6\%. 
Furthermore, it can be observed that given the same backbone, the model that applies HierCode achieves higher recognition accuracy than the one-hot counterparts.
Moreover, the application of HierCode in smaller-sized models yields a more significant compression ratio. 
For example, the parameters of ResNet-18, MobileNet v3 large, and MobileNet v3 small are compressed by 13.4\%, 49.0\% and 68.3\%, respectively. 
This suggests that the proposed HierCode facilitates the development of lightweight Chinese text recognition backbone. 

\begin{table}[t]
\small
\vspace{-4mm}
\caption{Comparison of HierCode and one-hot encoding in terms of the accuracy and parameters on different lightweight backbones. The `P-Cls', `P-Model', and `$\delta$' denote the parameters of the classification layer, the parameters of the whole model, and the compression ratio brought by HierCode.}
\vspace{1mm}
\label{Table 6}
\centering
\begin{threeparttable}
\begin{tabular}{llrrr}
\hline
\multirow{2}{*}{\textbf{Metric}} & \multirow{2}{*}{\textbf{Methods}} & \multicolumn{3}{c}{\textbf{Backbone}} \\
\cline{3-5}
& & ResNet-18 & MobileNet-L & MobileNet-S \\
\hline
\multirow{2}{*}{\textbf{Accuracy}} & One-hot/\% &97.64 & 95.57 & 92.13 \\
& HierCode/\% & \textbf{97.68} & \textbf{95.92} & \textbf{92.56} \\
\hline
\multirow{3}{*}{\textbf{P-Cls}} & One-hot/MB & 7.35 & 18.35 & 18.35 \\
 & HierCode/MB & \textbf{0.54} & \textbf{1.35} & \textbf{1.35} \\
 & $\delta$/\% & \textcolor{blue}{\textbf{92.60}} & \textcolor{blue}{\textbf{92.60}} & \textcolor{blue}{\textbf{92.60}} \\
\hline
\multirow{3}{*}{\textbf{P-Model}} & One-hot/MB & 50.97 & 34.70 & 24.89 \\
& HierCode/MB & \textbf{44.16} & \textbf{17.70} & \textbf{7.89} \\
& $\delta$/\% & \textcolor{blue}{\textbf{13.40}} & \textcolor{blue}{\textbf{49.00}} & \textcolor{blue}{\textbf{68.30}} \\
\hline
\end{tabular}
\end{threeparttable}
\end{table}

\subsection{Ablation Studies}
\label{Sec_Ablation}
In this subsection, we conduct a series of ablation experiments.
As illustrated in Tab.~\ref{Table_Ablation}, we establish the baseline by setting the structural features code length $L_\mathbf{S}$ to 4, the radical features code length $L_\mathbf{R}$ to 36, and the full binary tree depth $D$ to 5.
Since the maximum number of radicals $M$ in our experimental dataset does not exceed 9, $t$ can be calculated as 384 by Eq.~\ref{eq_t}. 

\begin{table}[h]
\small
\vspace{-5mm}
\centering
\caption{Ablation experiments on the design of radical features code length $L_\mathbf{R}$, structural features code length $L_\mathbf{S}$ and the full binary tree depth $D$.}
\vspace{1mm}
\begin{tabular}{lrrrrr}
\hline
      & $L_\mathbf{R}$    & $L_\mathbf{S}$    & $D$        & \textbf{AR}  & \textbf{CR} \\
\hline
Baseline & 36    & 4     & 5       & 91.98 & 92.33 \\
(a)   & 12    & 4     & 5       & 89.53 & 89.95 \\
(b)   & 24    & 4     & 5       & 91.22 & 91.57 \\
(c)   & 48    & 4     & 5      & 91.95 & 92.30 \\
\hline
(d)   & 36    & 8     & 5      & 91.83 & 92.24 \\
(e)   & 36    & 12    & 5     & 91.91 & 92.31 \\
\hline
(f)   & 36    & 4     & 6      & 91.99 & 92.36 \\
(g)   & 36    & 4     & 7      & 91.96 & 92.31 \\
\hline
\label{Table_Ablation}
\end{tabular}%
\vspace{-4mm}
\end{table}%


\textbf{The influence of $L_\mathbf{R}$.} 
Comparing the baseline with lines (a), (b), and (c), it is observed that shorter $L_\mathbf{R}$ constrains recognition performances. 
For instance, when $L_\mathbf{R}=12$, AR and CR are only 89.53\% and 89.95\%, respectively, which is 2.45\% and 2.38\% lower than the baseline. 
This is attributed to that a small code length not only fails to represent the full spectrum of radicals of Chinese characters, which has thousands of categories ($2^{12} = 4096 < 10k$), but also hampers the distinguishment of radical features in prototype learning.
With the increase of $L_\mathbf{R}$, the recognition performance gradually improves and approaches saturation.
Ultimately, we choose $L_\mathbf{R}=36$ as our standard setting as a trade-off for better performance and fewer model parameters. 

\textbf{The influence of $L_\mathbf{S}$.} 
By comparing the baseline with lines~(d) and (e), it can be seen that the recognition performance of $L_\mathbf{S}=4$ is comparable to that of $L_\mathbf{S}=8$ and $L_\mathbf{S}=12$. 
Given a limited number of character structures which is fixed to 10, the increase in $L_\mathbf{S}$ does not result in a notable improvement in recognition performance but brings additional model parameters. 
Therefore, we opt $L_\mathbf{S}=4$ as our standard setting. 

\textbf{The influence of $D$.} 
The depth of the full binary tree is closely related to the radical hierarchical decomposition of the characters.
If reducing $D$ to 4, we observe that numerous characters with large amounts of strokes and complex structures could not be completely decomposed, resulting in a sharp increase in the number of radicals and making it impossible to derive meaningful results. 
On the other hand, increasing $D$ to at least 5, enables the decomposition of most characters.
However, as demonstrated by experiments (f) and (g), there is no significant improvement in performance.
We select $D=5$ as the standard experimental setting. 

\textbf{The influence of HierCode's generation method.}
As described in Sec.~\ref{Sec_METH_2}, based on the hierarchical decomposition of Chinese characters, the proposed HierCode consists of structural features and radical features. 
Then, we maintain the overall composition of HierCode and replace the radical codes with randomly generated equal-length codes to experiment (b) in Tab.~\ref{Table_random}. 
By comparing settings (a) and (b) in Tab.~\ref{Table_random}, it can be seen that applying random radical codes slightly decreases the accuracy of ICDAR-char, AR, and CR of ICDAR-line by 0.22\%, 0.36\% and 0.37\%, respectively, which indicates that the pre-trained radical prototypes can better express the unique and robust features of radicals.
Notably, in cases where obtaining radical prototypes is unfeasible, e.g., in the zero-shot setting, the option of employing randomly generated radical codes becomes a viable alternative.
Furthermore, we randomly initialize a unique code for each character, whose length is equal to HierCode, to conduct (c). 
Note that the codes in (c) no longer have the hierarchical features of Chinese characters, which is equivalent to ordinary multi-hot coding. 
From the results in (c) of Tab.~\ref{Table_random}, it can be seen that randomness for the entire codes will significantly reduce the accuracy of ICDAR-char, AR, and CR of ICDAR-line by 3.09\%, 1.26\% and 1.31\%, respectively. 
In summary, the generation method of HierCode is meaningful, especially the hierarchical information of the Chinese characters, which is crucial to improving performance. 

\begin{table}[htbp]
\setlength\tabcolsep{4pt}
\small
\vspace{-2mm}
\centering
\caption{Ablation studies on the generation method of HierCode.}
\vspace{1mm}
\begin{tabular}{lrrrrr}
\hline

\multirow{2}[0]{*}{\textbf{Settings}} & {\textbf{ICDAR-char}} & & \multicolumn{2}{c}{\textbf{ICDAR-line}} \\
\cline{2-2}\cline{4-5}        & \textbf{Accuracy/\%} &        & \textbf{AR/\%}    & \textbf{CR/\%} \\
\hline
(a) HierCode~(Baseline) & 97.43 & & 91.98 & 92.33  \\
(b) Random radical code & 97.21 & & 91.62 & 91.96 \\
(c) Random entire code & 94.34 & & 90.72 & 91.02 \\
\hline
\end{tabular}%
\label{Table_random}%
\vspace{-5mm}
\end{table}%

\subsection{Visualization}
\begin{figure}[ht]
\centering
\vspace{-2mm}
\includegraphics[width=0.65\textwidth]{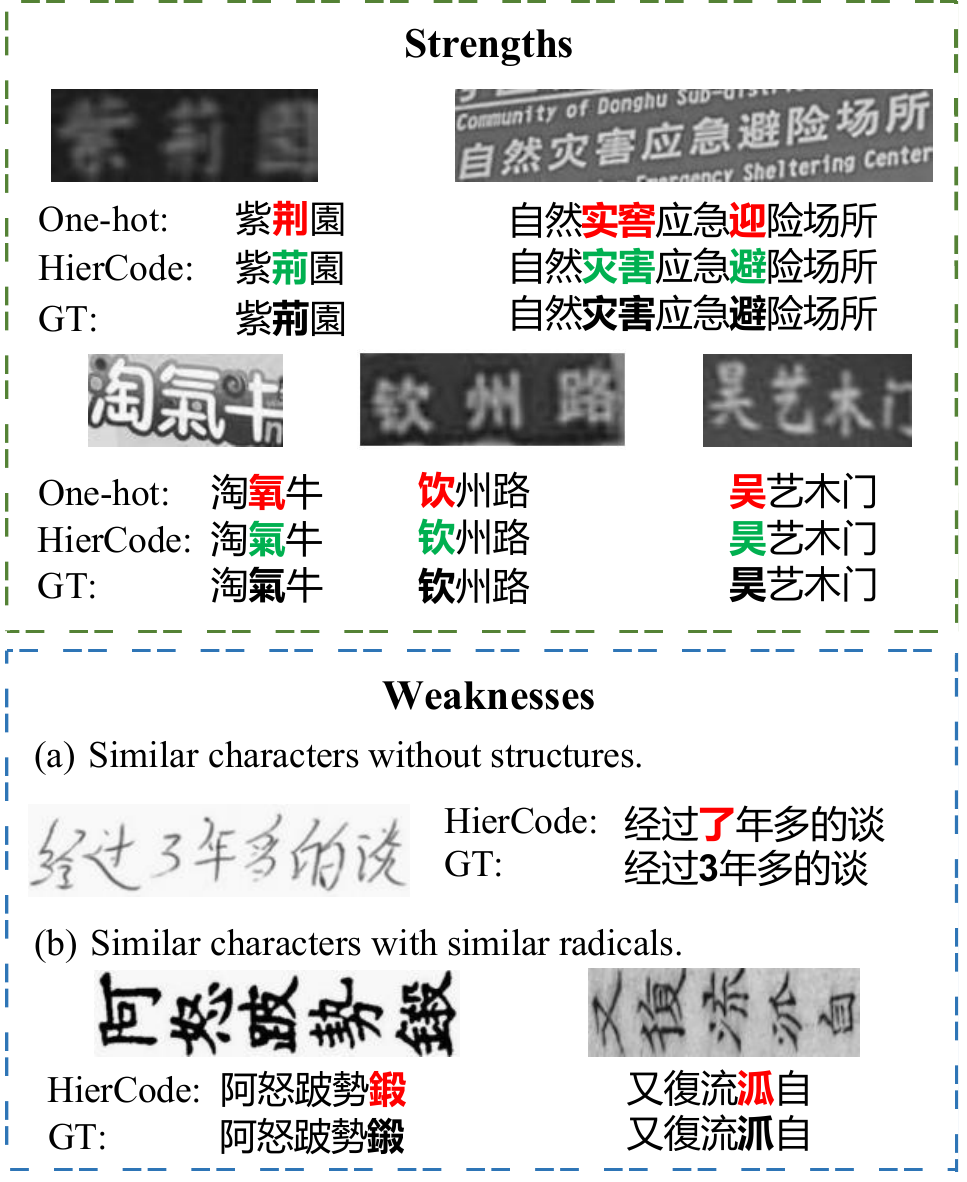}
\caption{Visualizations analysis of the strengths and weaknesses of HierCode. Correctly and incorrectly recognized characters are marked in \textcolor[RGB]{0,175,79}{\textbf{green}} and \textcolor{red}{\textbf{red}}, respectively. }
\label{figure_badcase}
\end{figure}

In this subsection, we present visualizations of various datasets to further analyze the strengths and weaknesses of the proposed HierCode. 
HierCode outperforms one-hot encoding in recognizing some similar characters. 
For instance, the two colored characters in the first subfigure in the `Strengths' part of Fig.~\ref{figure_badcase} have similar appearances and the same radicals, but different structures. 
Since their structural codes in HierCode are significantly different, our method can correctly recognize them, while the one-hot encoding method fails to do so. 

\subsection{Limitation \& Discussion }
HierCode suffers from some limitations.
For characters without structures, such as the single-radical Chinese characters, Arabic numerals, and Latin characters, HierCode degenerates into ordinary multi-hot encoding, resulting in the inability to distinguish these characters when they exhibit similar visual appearances.
Additionally, in cases where characters share similar radicals, HierCode may not effectively differentiate them. 
This issue represents the main limitation of radical-based methods and constitutes the primary focus of our future research endeavors. 


\section{Conclusion and Future Work}

This paper presents HierCode, an innovative lightweight hierarchical codebook designed for zero-shot Chinese text recognition. 
Through hierarchical encoding and prototype learning, HierCode assigns unique and informative representations to each Chinese character, capturing the inherent structural and effective radical features. 
Our method not only addresses the challenge of zero-shot character recognition, enabling accurate identification of characters unseen during training but also proves effective in line-level recognition tasks. 
HierCode’s use of multi-hot encoding significantly reduces the number of parameters required, resulting in a model that is both compact and capable of fast inference. 
The extensive experimentation across various datasets, including handwritten, scene, document, web, and ancient texts, demonstrates HierCode’s superior performance over traditional one-hot encoding methods and many state-of-the-art approaches.    
Future work will focus on improving the ability to recognize text with complex backgrounds and exploring more effective ways of generating radical feature codes to distinguish similar radicals.


\section*{}
\bibliography{egbib}

\end{document}